\pdfoutput=1

\documentclass[11pt]{article}
\usepackage{amsmath}
\usepackage{xspace}
\usepackage{amsfonts}
\usepackage{subcaption}
\usepackage{bm}
\usepackage{multirow}
\usepackage{booktabs} 
\usepackage{IEEEtrantools}
\usepackage{pifont}
\usepackage{color, colortbl}
\usepackage[inline]{enumitem}
\usepackage{listings}
\usepackage{ulem}
\usepackage{authblk}

\usepackage[final]{acl}

\usepackage{times}
\usepackage{latexsym}

\usepackage[T1]{fontenc}

\usepackage[utf8]{inputenc}

\usepackage{microtype}

\usepackage{inconsolata}

\usepackage{graphicx}
\definecolor{Highlight}{rgb}{0.89,0.89,0.94}
\newcommand{\chl}{\cellcolor{Highlight}}
\newcommand{\mymodel}{\textsc{LoVer}\xspace}

\title{Logic-Regularized Verifier Elicits Reasoning from LLMs}

\author[*1,2\#]{\bf Xinyu Wang}
\author[*2]{\bf Changzhi Sun}
\author[1,2\#]{\bf Lian Cheng}
\author[1]{\bf Yuanbin Wu}
\author[2]{\\ \bf Dell Zhang}
\author[1$\dag$]{\bf Xiaoling Wang}
\author[2$\dag$]{\bf Xuelong Li}
\affil[1]{Department of Computer Science and Technology, East China Normal University}
\affil[2]{Institute of Artificial Intelligence (TeleAI), China Telecom}
\affil[ ]{\tt\{xinyu\_wang@stu, ybwu@cs, xlwang@cs\}.ecnu.edu.cn}
\affil[ ]{\tt \{czsun\}@chinatelecom.cn }
\affil[ ]{\tt \{dell.z, xuelong\_li\}@ieee.org}

\begin{document}

\maketitle

\begingroup
\renewcommand\thefootnote{*}
\footnotetext{Equal contribution.}
\renewcommand\thefootnote{\#}
\footnotetext{Work done while this author was an intern at TeleAI.}
\renewcommand\thefootnote{$\dag$}
\footnotetext{Corresponding authors.}
\endgroup

\begin{abstract}


Verifiers are crucial components for enhancing modern LLMs' reasoning capability.
Typical verifiers require resource-intensive supervised dataset construction, which is costly and faces limitations in data diversity.
In this paper,  we propose \mymodel, an unsupervised verifier regularized by logical rules. 
\mymodel treats the verifier as a binary latent variable, utilizing internal activations and enforcing three logical constraints on multiple reasoning paths: negation consistency, intra-group consistency, and inter-group consistency (grouped by the final answer). 
By incorporating logical rules as priors, \mymodel can leverage unlabeled examples and is directly compatible with any off-the-shelf LLMs.
Experiments on $10$ datasets demonstrate that \mymodel significantly outperforms unsupervised baselines, achieving performance comparable to the supervised verifier (reaching its 95\% level on average).
The source code is publicly available at \url{https://github.com/wangxinyufighting/llm-lover}.


\end{abstract}

\section{Introduction}
\label{sec:intro}
Verifiers guide LLMs by providing feedback to optimize their parameters (RL scaling) or outputs (inference scaling), 
which greatly enhances models' reasoning capabilities~\cite{ouyang2022training,snell2024scaling}.
Verifiers are usually trained with supervised learning~\cite{cobbe2021training, yu-etal-2024-ovm}, where they learn to classify reasoning outputs as true or false based on labeled data.
It presents two challenges: 
1) The verifier relies heavily on labeled data for training, which can be expensive to collect (particularly in specialized or complex domains).
For example, annotating a single Olympiad-level problem typically takes a significant amount of time, and adding fine-grained step-level process annotations ~\cite{lightman2023let} further increases the workload;
2) Relying on expert annotations may result in a lack of diversity in the solutions~\cite{basile-etal-2021-need, xu-etal-2024-leveraging}, as annotators may favor familiar reasoning methods while overlooking equally valid but less intuitive ones.
For example, when annotating geometric problems, annotators may prefer the standard coordinate method and down-vote the less  obvious geometric observation.
While one can improve the supervision process in various aspects~\cite{yang-etal-2019-predicting} (e.g., more experts with diverse mathematical backgrounds and education experiences), 
LLMs themselves already compact large amounts of knowledge and abilities to sample diverse generations~\cite{minaee2024largelanguagemodelssurvey, xu2024survey}, a natural question is \normalem{\emph{whether we could build verifiers without the supervision process?}}


\begin{table*}[tb!]
    \centering
    \footnotesize
    \begin{tabular}{lcccccc}
        \toprule
            \textbf{Verifier}            & \textbf{Paradigm}       & \textbf{Prior}   &     \textbf{Annotation}            & \textbf{Input}    &   \textbf{Model} &   \textbf{Scenario} \\
        \midrule
        \vspace{1pt}
        \citet{NEURIPS2024_7a8e7fd2}     & Unsup. &  Heuristics & -             &  Probabilities &  -  & General\\
        \vspace{1pt}
        \citet{burns2023discovering}      &      Unsup.     & Logic rule  & - & Hidden states & Linear &  Yes-No\\
        \vspace{1pt}
        \citet{cobbe2021training} &    Sup.        &  -         &     Outcome-based                   & Text & Decoder-only & General \\
        \vspace{1pt}
        \citet{lightman2023let}  & Sup. &    -       &       Process-based        &     Text     &  Decoder-only &  General \\
        \vspace{1pt}
        \mymodel & Unsup. &    Logic rule        & -             &    Hidden states        & MLP & General\\
        \bottomrule
    \end{tabular}
    \vspace{-5pt}
    \caption{ Comparison between existing verifiers.
   \textbf{Paradigm}: The verifier is trained using a supervised (Sup.) or unsupervised (Unsup.) learning paradigm.
   \textbf{Prior}: The prior knowledge used in the verifier.
   \textbf{Annotation}: The type of annotation data.``Outcome-based'' is solution-level annotation. ``Process-based'' is step-level annotation.
   \textbf{Input}: The input data type of the verifier. 
   \textbf{Model}: The model architecture.
   \textbf{Scenario}: Reasoning scenarios suitable for the verifier. ``General'' typically refers to reasoning problems that have a correct answer. ``Yes-No'' indicates that the answer to the question is either Yes or No.
    }
    
    \label{tab:abc}
    \vspace{-15pt}
    
\end{table*}


To address these challenges, recent research has focused on unsupervised verifiers to uncover the intrinsic reasoning capabilities of LLMs.
Typical works include: 1) CoT-Decoding~\cite{NEURIPS2024_7a8e7fd2}, which proposes a heuristic rule-based verifier by observing the probabilities of the outputs of LLMs. 
It selects the correct reasoning path based on the probability difference between the top and secondary tokens in the answer span. 
In experiments, we observed that CoT-Decoding is sensitive to the backbone LLM choice. 
For example, when using \texttt{llama-7b} on the GSM8K dataset, CoT-Decoding is 4.8\% lower than the majority voting strategy.
2) CCS~\cite{burns2023discovering} introduces an unsupervised verifier, which is essentially a linear probe optimized through logical consistency loss. 
Unfortunately, CCS can only address Yes-No questions and struggles to scale to general reasoning tasks.
A practical verifier should have fewer limitations on its target problems, enabling it to handle a broader range of reasoning scenarios and provide more flexibility in real-world applications

In this paper, we propose a principled framework \mymodel, an unsupervised probabilistic verifier regularized by logical rules.
For each reasoning path, we search for the implicit, internal ``beliefs'' or ``knowledge'' learned by the LLM to infer the truth value of the reasoning.
\mymodel begins by generating contrastive assertions through incorporating text templates. 
It then takes the internal activations of these assertions from the LLM as inputs and produces a binary latent variable to indicate the truth value.
Furthermore, \mymodel incorporates three logical constraints including negation consistency, intra-group consistency, and inter-group consistency (with multiple reasoning paths grouped by the final answer).
To bridge the gap between discrete logical rules and continuous neural networks, we propose corresponding soft probabilistic objectives that support differentiable training.
Our contributions are summarized as follows:
\begin{itemize}[noitemsep, leftmargin=1pt]
    \item We propose \mymodel, a scalable and principled framework for verifying the truth value of reasoning paths, leveraging intrisic knowledge learned by the LLM and regularized by logical rules.
    Additionally, \mymodel is fully compatible with any off-the-shelf LLMs.
    \item To combine discrete logical rules with neural networks, we propose soft probabilistic objectives that enable \mymodel to be trained end-to-end, improving its scalability and performance.
    \item Our extensive experiments across diverse datasets, including mathematical reasoning, common sense reasoning, and various backbones, demonstrate the effectiveness of the proposed method.
\end{itemize}

\section{Approach}
\label{sec:approach}
\label{sec:approach}
In this section, we present the proposed \mymodel, an unsupervised verifier designed to reason over the internal activations of LLMs.

\begin{figure*}
    \centering
    \includegraphics[width=1\linewidth]{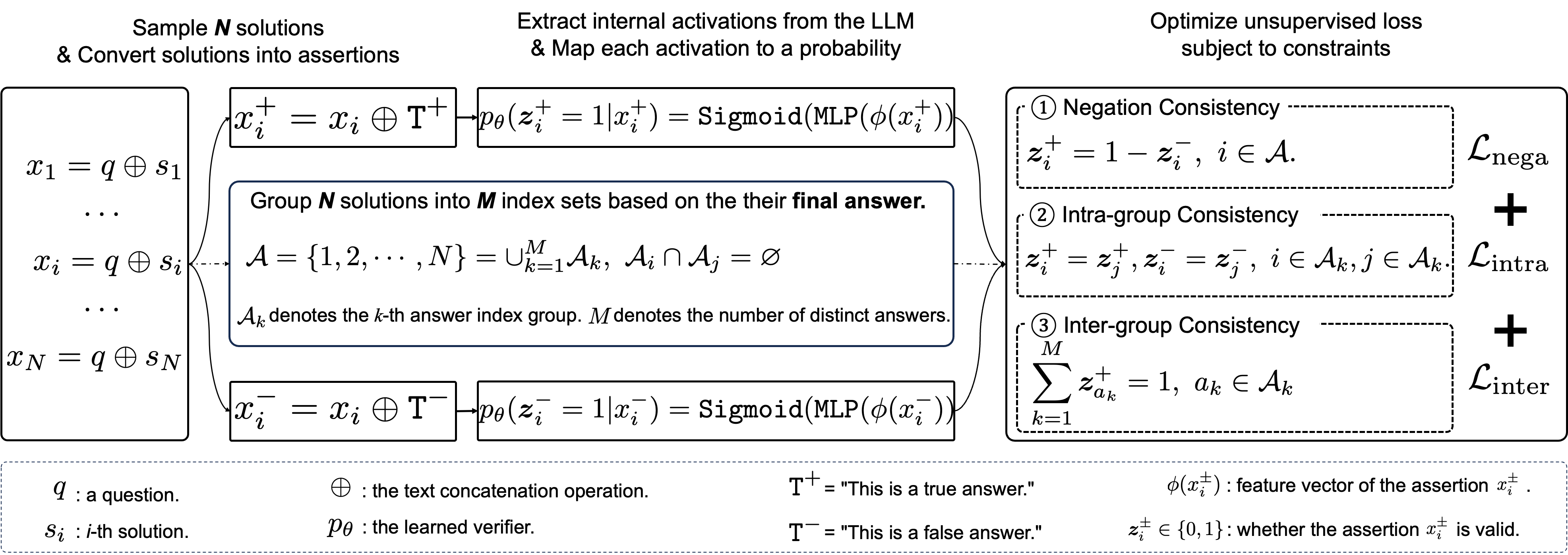}
    \caption{
        An illustration of our proposed \mymodel.
        For any question $q$, we create $x_i$ by combining $q$ with the $i$-th solution from $N$ solutions. 
        We form $x^+_i$ and $x^-_i$ by adding "This is a true/false answer." to $x_i$, respectively. 
        Choosing the correct solution involves determining which assertion, $x^+_i$ or $x^-_i$, is correct.
        The hidden states of LLMs are used to represent $x^+_i$ and $x^-_i$, which are then input into \mymodel to predict the correctness probability of each assertion.
        We extract the final answer from each solution and group assertions with identical answers together.
        These assertions follow three natural logical constraints that guide \mymodel's unsupervised training. 
        \textbf{Negation Consistency} ensures that only one of $x^+_i$ or $x^-_i$ is correct.
        \textbf{Intra-group Consistency} requires that assertions in the same group have equal correctness probabilities. 
        \textbf{Inter-group Consistency} ensures that only one group's $x^+$ assertion is correct across all groups.
    }
    \label{fig:overview}
    
\end{figure*}

\paragraph{Task Definition}
Given an LLM and an input question $q$, we first generate $N$ complete solutions $\{s_i\}_{i=1}^N$, with each $s_i$ representing a CoT path (Sec. ~\ref{sec:best-of-n}).
We then select the best solution based on a learned verifier.
For each solution $s_i$, we define $x_i = q \oplus s_i, x_i^+ = x_i \oplus \mathtt{T}^+, x_i^- = x_i \oplus \mathtt{T}^-$,
where $\oplus$ denotes the text concatenation, and $\mathtt{T}^+, \mathtt{T}^-$ are text templates.
Given $N$ reasoning paths, we group them into $M$ sets ($M \leq N$) based on the \normalem{\emph{final answer}} (extracted through rules from the answer token). 
$\mathcal{A}$ represents the index set from $1$ to $N$, and $\mathcal{A}_k$ denotes the  $k$-th group, with $\mathcal{A} = \cup_{k=1}^M \mathcal{A}_k$.
The verifier models a probabilistic distribution $p_\theta(\bm{z} | x)$, where $x \in \cup_{i=1}^N \{ x_i^+, x_i^-\}$ and $\bm{z} \in \{0, 1\}$ is a binary latent variable indicating whether the natural language statement $x$ is valid.
In this paper, bold letters indicate variables.

Inspired by CoT-Decoding~\cite{NEURIPS2024_7a8e7fd2} and CCS~\cite{burns2023discovering}, 
to find the correct answer, we first augment each reasoning path to derive its correct and incorrect assertions, and then treat the truth values of the assertions as binary latent variables.
On the one hand, we leverage the internal activations of the LLM as input, enabling better utilization of the model's intrinsic knowledge.
On the other hand, the logical constraints provide implicit supervision signals to update the verifier, significantly reducing the need for human supervision.

Next, we first introduce LLM decoding strategy (Sec.~\ref{sec:best-of-n}) and how to obtain contrastive assertions (Sec.~\ref{sec:augment}).
Then we detail the latent verifier model (Sec.~\ref{sec:verifier}) and describe the logical constraints imposed on the latent variables (Sec.~\ref{sec:logic}).
Finally, we present the training and inference procedure (Sec.~\ref{sec:training_and_inference}).
Fig.~\ref{fig:overview} shows an overview of our method.

\subsection{LLM Decoding Strategy}
\label{sec:best-of-n}
Given an input question $q$ and a typical decode-only LLM, there are various strategies to decode $N$ solutions, such as beam search, nucleus sampling, and others.
In this work, we follow the CoT-Decoding~\cite{NEURIPS2024_7a8e7fd2}.
Specifically, we keep the top $N$ tokens with the highest probabilities at decoding step $0$, and then continue with greedy decoding for each token, ultimately producing $N$ solutions. 
Compared to other strategies, this method is more likely to produce a natural CoT reasoning path and does not rely on complex prompt engineering~\cite{NEURIPS2024_7a8e7fd2}.
In the experiments, we also study the impact of different decoding strategies(Table~\ref{tab:result_diff_sampling}).

\subsection{Contrastive Assertions}
\label{sec:augment}
For each $x_i = q \oplus s_i$, we construct each contrastive assertions by appending the text templates $\mathtt{T}^+$ and $ \mathbb{T}^-$.
Formally, this is denoted as $x_i^+ = x_i \oplus \mathtt{T}^+$ and $x_i^- = x_i \oplus \mathtt{T}^-$.
In this paper, we adopt $\mathtt{T}^+=$ \texttt{``This is a true answer.''} and $\mathtt{T}^-=$\texttt{``This is a false answer.''}.
Importantly, rather than directly considering each reasoning path $x_i$, we introduce contrastive assertions $x_i^+$ and $x_i^-$, which help elicit the internal ``beliefs'' or ``knowledge'' learned by the model~\cite{burns2023discovering}.

\subsection{Latent Verifier Model}
\label{sec:verifier}
For each natural language assertion $x \in \cup_{i=1}^N \{ x_i^+, x_i^-\}$, we  first compute the feature vector of $x$, denoted as 
$\phi(x)$,
\footnote{The default is the hidden representation of the last token in the middle layer, and we also explore other options. For details, please refer to Table~\ref{tab:result_diff_layers}.}
then pass it through a randomly initialized MLP, and finally map it to a probability value using the sigmoid function. 
Formally, we define the probabilistic distribution of verifier $p_\theta(\bm{z} | x)$ where $\bm{z} \in \{0, 1\}$ is a binary latent variable indicating whether the natural language statement $x$ is valid.
For simplicity, we use $p_\theta(\bm{z})$ to represent $p_\theta(\bm{z} = 1 | x)$:
\begin{IEEEeqnarray*}{c}
p_\theta(\bm{z}) = p_\theta(\bm{z} = 1| x) = \mathtt{Sigmoid}(\mathtt{MLP}(\phi(x)).
\end{IEEEeqnarray*}
Importantly, \mymodel does not modify the weights of the LLM and it does not use labels.

\subsection{Logical Constraints}
\label{sec:logic}
After introducing the binary latent variables $\cup_{i=1}^N \{\bm{z}^+, \bm{z}^+\}$, we observe that certain natural logical consistencies between them should be satisfied.
Let us look at three such logical consistency requirements.

\paragraph{Negation Consistency}
Given the contrastive assertions $x_i^+$ and $ x_i^-$, their corresponding binary latent variables $\bm{z}_i^+$ and $\bm{z}_i^-$  should satisfy negation consistency:
\begin{IEEEeqnarray*}{c}
\bm{z}_i^+ = 1 -  \bm{z}_i^-, i \in  \mathcal{A}.
\end{IEEEeqnarray*}
To this end, we relax the logic with soft probability ~\cite{chen2022loren,burns2023discovering} for differentiability in training and regularization of binary latent variables.
Inspired by CCS~\cite{burns2023discovering}, we aim for the contrastive assertions $x_i^+$ and $x_i^-$ to satisfy the following:
1).the sum of their probabilities equals $1$ (probability normalization);
2).their probabilities differ significantly (the law of excluded middle).
\begin{IEEEeqnarray*}{ll}
\mathcal{L}_\mathrm{sum} ~ &=  \sum_{i=1}^N  \left [p_\theta(\bm{z}_i^+)+ p_\theta(\bm{z}_i^-) - 1 \right ]^2, \\
\mathcal{L}_\mathrm{diff} ~ &= \sum_{i=1}^N  \min \left \{ p_\theta(\bm{z}_i^+), p_\theta(\bm{z}_i^-) \right \}^2, \\
\mathcal{L}_\mathrm{nega} ~ &= \mathcal{L}_\mathrm{sum} + \mathcal{L}_\mathrm{diff}.
\end{IEEEeqnarray*}
Note that both losses are necessary; using either one alone leads to a degenerate solution~\cite{burns2023discovering}.

\paragraph{Intra-group Consistency}
For each group $\mathcal{A}_k$ of reasoning paths, they share the same answer, though their reasoning processes may differ. 
Overall, we expect the corresponding binary latent variables to satisfy intra-group consistency, i.e.,
\begin{IEEEeqnarray*}{l}
\bm{z}_i^+ = \bm{z}_j^+,  \bm{z}_i^- = \bm{z}_j^-, ~ ~ ~ i \in  \mathcal{A}_k, j \in \mathcal{A}_k. 
\end{IEEEeqnarray*}
To achieve this goal, we use a simple squared loss:
\begin{IEEEeqnarray*}{ll}
\mathcal{L}_\mathrm{intra}^+ &= \sum_{k=1}^M \sum_{i \in \mathcal{A}_k, j \in \mathcal{A}_k} \left [ p_\theta(\bm{z}_i^+) - p_\theta(\bm{z}_j^+) \right ]^2, \\
\mathcal{L}_\mathrm{intra}^- &= \sum_{k=1}^M \sum_{i \in \mathcal{A}_k, j \in \mathcal{A}_k}  \left [p_\theta(\bm{z}_i^-) - p_\theta(\bm{z}_j^-) \right ]^2, \\
\mathcal{L}_\mathrm{intra} &= \mathcal{L}_\mathrm{intra}^+ + \mathcal{L}_\mathrm{intra}^-.
\end{IEEEeqnarray*}

\paragraph{Inter-group Consistency}
Among the $N$ reasoning paths, there are $M$ distinct answers. 
We assume that the LLM's capabilities are sufficiently strong to ensure the presence of a correct answer.
We examine the GSM8k dataset and find that when $N = 10$, the $P @ 10$ accuracy of \texttt{qwen-2.5} can reach $91.43\%$.
This confirms the validity of the above assumption.
Specifically, for each group $\mathcal{A}_k$, we randomly select an $a_k$ and hope that its corresponding binary latent variable satisfies:
\begin{IEEEeqnarray}{l}
\label{eq:inter}
\sum_{k=1}^M \bm{z}_{a_k}^+ = 1, ~ ~ ~ a_k \in \mathcal{A}_k.
\end{IEEEeqnarray}
To achieve this inter-group consistency, we propose a soft probability solution.
\begin{IEEEeqnarray*}{ll}
\mathcal{L}_\mathrm{inter}^\mathrm{sum} &= \left [ \sum_{k=1}^M p_\theta(\bm{z}_{a_k}^+) - 1 \right ]^2.
\end{IEEEeqnarray*}
However, in the experiments, we observe that if only the loss $\mathcal{L}_\mathrm{inter}^\mathrm{sum}$ is used, the $M$ probabilities  $p_\theta(\bm{z}_{a_k}^+) $ tend to become uniform.
To address this issue, we propose an entropy regularization.
We introduce a probability distribution $\hat{p}$ which defines over the $M$ variables $\{ \bm{z}_{a_k}^+ \}_{k=1}^M$.
\begin{IEEEeqnarray*}{ll}
\mathcal{L}_\mathrm{inter}^\mathrm{h} &= \mathcal{H}\left [ \hat{p}(\cdot) \right ], ~ ~ ~ 
\hat{p}(\bm{z}_{a_k}^+) = \frac{p_\theta(\bm{z}_{a_k}^+)}{\sum_{i=1}^M p_\theta(\bm{z}_{a_i}^+)},\\
\mathcal{L}_\mathrm{inter} &= \mathcal{L}_\mathrm{inter}^\mathrm{sum} + \mathcal{L}_\mathrm{inter}^\mathrm{h},
\end{IEEEeqnarray*}
where $\mathcal{H}$ denote the entropy function.
In addition, we also explore a soft logic-based solution, which is encapsulated in Appendix~\ref{sec:logic_based}.

\subsection{Training and Inference}
\label{sec:training_and_inference}
\paragraph{Training}
\label{sec:trianing}
The final loss function is the sum of three losses mentioned above, which is defined as:
\begin{IEEEeqnarray*}{c}
\mathcal{L} = \mathcal{L}_\mathrm{nega} + \mathcal{L}_\mathrm{intra}  + \mathcal{L}_\mathrm{inter} .
\end{IEEEeqnarray*}
\mymodel is structured as a MLP with 2 hidden layers, and we select ReLU as the activation function.
We use AdamW\cite{loshchilov2017decoupled} as optimizer (\texttt{weight\_decay} = $0.01$),
and set learning rate to $1\times e^{-5}$.

\paragraph{Inference}
\label{sec:inference}
Given an input question $q$, we first decode $N$ candidate solutions (Sec. \ref{sec:best-of-n}) to obtain $\{x_i\}_{i=1}^N$.
For each $x_i$, we generate contrastive assertions $x_i^+$ and $x_i^-$ of $s_i$  (Sec.~\ref{sec:augment}) and compute corresponding probability $p_\theta(\bm{z}_i^+)$ and $p_\theta(\bm{z}_i^-)$  based on the latent verifier model (Sec.~\ref{sec:verifier}).
Both $p_\theta(\bm{z}_i^+)$ and  $1 - p_\theta(\bm{z}_i^-)$ should represent the probability that the $x_i$ is correct.
we consequently take the average of these~\cite{burns2023discovering}: 
\begin{IEEEeqnarray*}{c}
p_\theta(\bm{z}_i) =\frac{1}{2}\left [ p_\theta(\bm{z}_i^+) + (1 - p_\theta(\bm{z}_i^-))\right ].
\end{IEEEeqnarray*}
Then we group them into $M$ sets based in the final answer (extracted through rules from answer token).
For each group $\mathcal{A}_k$, we compute the group score $g_k$ using two strategies: \normalem{\emph{max}}  and \normalem{\emph{sum}}.
The \normalem{\emph{max}} strategy computes the group score by selecting the maximum $p_\theta$ within the group: $g_k = {\max}_{i \in \mathcal{A}_k} p_\theta(\bm{z}_i)$.
The \normalem{\emph{sum}} strategy computes the group score by summing up all $p_\theta$ within the group: $g_k = {\sum}_{i \in \mathcal{A}_k} p_\theta(\bm{z}_i)$.
Finally, we select the answer with the highest group score $g_k$ among the $M$ groups.
Appendix \ref{sec:pseudo} provides PyTorch-style pseudocode for the inference procedure.

\section{Experiments}
\label{sec:exp}
\begin{table*}[h]
\centering
\begin{tabular}{clcccc}
\toprule
\multirow{2}{*}{\textbf{LLM}}  & \multirow{2}{*}{\begin{tabular}[c]{@{}l@{}}\textbf{Dataset} $\rightarrow$\\ \textbf{Verifier} $\downarrow$\end{tabular}} & \multicolumn{2}{c}{\textbf{Mathematics}} & \multicolumn{2}{c}{\textbf{Open-Domain Knowl.}} \\ 
                            &                                                                             & \textbf{GSM8K}           & \textbf{iGSM}       & \textbf{HotpotQA}        & \textbf{MMLU-P}      \\ \midrule
\multirow{6}{*}{\texttt{mistral-7b}} & Supervised                                                         & 61.18           &44.50             & 82.96         & 41.93      
\\
\cmidrule{2-6}
                            & Greedy                                                                      & 16.15           & 19.00         & 70.74           &37.65   \\
                            & Majority Voting                                                             & 43.82           &34.50          & 76.86           &41.07         \\
                            & CoT-Decoding (max)                                                          & 36.62           &12.75          & 69.87           &38.04          \\
                            & CoT-Decoding (sum)                                                          & 47.76           &36.75          & 76.86           &41.43          \\
                            &\mymodel (max)                                                               & 50.41 \chl               &28.75\chl      & 74.24\chl       &40.36\chl    \\
                            &\mymodel (sum)                                                               & \textbf{53.14} \chl      &\textbf{41.25}\chl      & \textbf{77.95}\chl       &\textbf{41.82}\chl \\
                             \midrule
\multirow{6}{*}{\texttt{llama-8b}}   & Supervised                                                         & 83.24           &43.75              & 82.96            & 55.86      
\\
\cmidrule{2-6}
                            & Greedy                                                                      & 36.09           &35.00          & 77.29           &45.67              \\
                            & Majority Voting                                                             & 75.96           &39.00          & 81.00           &50.58       \\
                            & CoT-Decoding (max)                                                          & 38.51          &23.25           & 77.95           &45.92     \\
                            & CoT-Decoding (sum)                                                          & 71.11           &40.50          & 78.16           &54.61          \\
                            & \mymodel (max)                                                              & 70.60\chl       &39.00\chl      & 81.04\chl       &52.65\chl               \\
                            & \mymodel (sum)                                                              & \textbf{79.97}\chl &\textbf{42.50}\chl      & \textbf{83.80}\chl       &\textbf{55.61}\chl               \\
                            \midrule
\multirow{6}{*}{\texttt{qwen-7b}}    & Supervised                                                         & 92.11           &54.25              & 83.62           & 61.13      
\\
\cmidrule{2-6}
                            & Greedy                                                                      & 61.71           &46.50          & 78.82           &54.93              \\
                            & Majority Voting                                                             & 89.46           &51.00          & \textbf{83.41}           &57.00          \\
                            & CoT-Decoding (max)                                                          & 63.84           &31.50          & 81.66           &54.22          \\
                            & CoT-Decoding (sum)                                                          & 89.76           &51.00          & 82.09           &58.81          \\
                            & \mymodel (max)                                                              & 83.47\chl       &48.25\chl      & 81.41\chl       &59.28\chl               \\
                            & \mymodel (sum)                                                              & \textbf{91.43}\chl&\textbf{52.00}\chl      & 82.75\chl       &\textbf{61.24}\chl               \\
                            \bottomrule
\end{tabular}
\caption{
  \label{tab:main_results_all}
    The overall experimental results of \mymodel and other baselines on the four datasets.
    Accuracy is utilized to measure the performance.
    The best results of each setting are in bold. 
    MMLU-P stands for "MMLU-Pro" dataset.
}
\end{table*}



\paragraph{Datasets}
We conduct experiments on datasets of both mathematical and open-domain knowledge reasoning.
For mathematical reasoning, we use the Grade-school math problems, GSM8K \cite{cobbe2021training} and more challenging iGSM dataset \cite{YXLA2024-gsm1}.  
For open-domain knowledge reasoning, we use HotpotQA\cite{DBLP:conf/emnlp/Yang0ZBCSM18} and MMLU-Pro\cite{DBLP:journals/corr/abs-2406-01574}.
Furthermore, to evaluate the out-of-distribution (OOD) generalization of \mymodel, we employ Boolean Expressions, Web of Lies, Object Counting, Navigate, Multi-Step Arithmetic and Causal Judgement from BIG-Bench Hard\cite{DBLP:conf/acl/SuzgunSSGTCCLCZ23}.
The details of datasets are provided in Appendix~\ref{sec:datasets}.

\paragraph{Evaluation}
We evaluate accuracy by \normalem{\emph{strictly matching}} the final answer from the response with the ground truth answer. 

\paragraph{Baselines.}
We test three open-source LLMs: 
\texttt{llama-8b}, \texttt{mistral-7b}  and \texttt{qwen} with different scales, ranging from 0.5B, 1.5B, 3B, 7B, and 32B.
\footnote{
Specific versions are 
\texttt{llama-3.1-8b-instruct} \cite{DBLP:journals/corr/abs-2407-21783}
, \texttt{mistral-7b-instruct-v0.3} \cite{DBLP:journals/corr/abs-2310-06825}
, \texttt{qwen2.5-instruct} \cite{qwen2025qwen25technicalreport}. 
}
We compare \mymodel against following methods:
\begin{itemize}[noitemsep, leftmargin=1pt]
    \item ``Greedy'' decoding selects the most probable token at each step.
    \item ``Majority Voting'' decodes multiple outputs and select the optimal answer by voting \cite{DBLP:conf/nips/LewkowyczADDMRS22, DBLP:conf/iclr/0002WSLCNCZ23}. \footnote{The default decoding strategy is described in Sec.~\ref{sec:best-of-n}.}
    \item ``CoT-Decoding'' \cite{NEURIPS2024_7a8e7fd2} selects correct reasoning paths based on answer confidence (probability disparity between the top and secondary tokens in answer spans). 
    \item  ``Supervised'' is the supervised \mymodel. 
    It is trained using gold label data, with the training objective being the standard binary cross-entropy loss.
    Theoretically, this is the ceiling of \mymodel.
\end{itemize}

\subsection{Main Results}

\normalem{\emph{\mymodel effectively enhances reasoning abilities across models and reasoning types.}}
As shown in Table~\ref{tab:main_results_all}, \mymodel (sum) achieves the highest accuracy in all scenarios.
\mymodel (max) outperforms CoT-Decoding in $40\%$ of cases and matches the average accuracy of Majority Voting.

\mymodel (sum) achieves an average absolute accuracy gain of $3.3\%$ over Majority Voting.
Unlike Majority Voting, which relies solely on the frequency of answers, \mymodel not only considers answer counts but also leverages the internal knowledge of LLMs. 
Driven by logical constraints, \mymodel can more effectively utilize the LLMs' internal knowledge to score the correctness of assertions. 
Thus, \mymodel represents an optimized and weighted voting method.
Compared to CoT-Decoding, \mymodel shows an average absolute accuracy gain of $2.9\%$.
\mymodel focuses on the correctness of the solution itself, rather than emphasizing the format of the solution as in CoT-Decoding.
CoT-Decoding aims to elicit reasoning paths with CoT processes, leading to significant accuracy gains on weaker models
(those unable to autonomously generate CoT solutions without CoT prompting) 
but limited improvements on stronger models.
As a result, compared to CoT-Decoding, \mymodel is less affected by the underlying capabilities of the LLM.
\mymodel (sum) consistently outperforms \mymodel (max), demonstrating the effectiveness of the \normalem{\emph{sum}} strategy and highlighting the importance of the frequency of answers.
CoT-Decoding (max) achieves an average accuracy similar to Greedy, indicating that relying solely on the probability with answer tokens is insufficient.

\subsection{Ablation Studies}
\subsubsection{The effect of different logic constraints}
\label{sec:res_diff_loss}

\normalem{\emph{Incorporating logical constraints can significantly enhance \mymodel's performance.}}
Table~\ref{tab:result_diff_loss} reveals that the exclusion of the $\mathcal{L}_\mathrm{inter}$ led to the most significant drop in reasoning accuracy, indicating its crucial role in enhancing model performance.
Without $\mathcal{L}_\mathrm{inter}$, \mymodel tends to optimize towards assigning a score of 1 to all $x_i^+$ and 0 to all $x_i^-$. 

\begin{table}[h]
\centering
\begin{tabular}{l|cc|cc}
\hline
\multicolumn{1}{c|}{\multirow{2}{*}{Loss}} & \multicolumn{2}{c|}{\textbf{GSM8K}} & \multicolumn{2}{c}{\textbf{MMLU-Pro}} \\
\multicolumn{1}{c|}{}                      & \textbf{sum}     & \textbf{max}     & \textbf{sum}      & \textbf{max}      \\ \hline
\mymodel                                          & \textbf{53.14}   & \textbf{50.41}   & \textbf{41.82}    & \textbf{40.36}    \\
w/o $\mathcal{L}_\mathrm{nega}$                                 & 51.78            & 34.79            & 40.07             & 37.37             \\
w/o $\mathcal{L}_\mathrm{intra}$                                  & 52.76            & 44.20            & 41.75             & 40.00             \\
w/o $\mathcal{L}_\mathrm{inter}$                                  & 49.96            & 38.13            & 38.15             & 35.44             \\ \hline
\end{tabular}
  \caption{
  \label{tab:result_diff_loss}
    Accuracy of \mymodel on GSM8K and MMLU-Pro using different setting of logic constraints over \texttt{mistral-7b}.
  }
\end{table}
In this scenario, \mymodel loses the ability to discern the correctness of assertions.
Removing $\mathcal{L}_\mathrm{nega}$ also results in a noticeable decrease in accuracy. 
Without $\mathcal{L}_\mathrm{nega}$, the premise that each assertion has only one correctness label cannot be satisfied. 
As a result, \mymodel tends to optimize towards assigning identical scores to both  $x_i^+$ and  $x_i^-$.
The removal of $\mathcal{L}_\mathrm{intra}$ has a minimal impact on reasoning accuracy.
$\mathcal{L}_\mathrm{intra}$  enforces consistency in correctness probabilities for solutions with the same final answer. 
However, solution correctness depends not only on the final answer but also on the problem-solving process, which may contain errors even if the answer is correct.
Enforcing the consistency of correctness probabilities solely based on the same final answer may have limitations.

Logical constraints have a more substantial impact on \mymodel (max) than on \mymodel (sum). Insufficient constraints hinder \mymodel (max)'s ability to accurately assess assertion validity, while \mymodel (sum) mitigates this by incorporating answer frequency, reducing sensitivity to constraint variations.

\subsubsection{The effect of decoding strategies} 

\normalem{\emph{\mymodel achieves hightest reasoning accuracy combined with different decoding strategies.}}
Table~\ref{tab:result_diff_sampling} shows that all methods achieve their highest accuracy under the natural CoT-Decoding, outperforming temperature sampling and beam search sampling by an average of $19.5\%$ and $22.3\%$.

\begin{table}[h]
\begin{tabular}{l|ccc}
\toprule
\textbf{Decoding Strategies} &\textbf{Nat} & \textbf{Temp} & \textbf{Beam} \\ \midrule
Majority Voting         & 43.82       & 27.07       & 24.79       \\
CoT-Decoding & 47.76       & 28.05       & 24.94       \\
\mymodel     & \textbf{53.14}       & \textbf{31.31}       & \textbf{28.20}       \\ \bottomrule
\end{tabular}
\caption{
  \label{tab:result_diff_sampling}
    Accuracy of \mymodel on GSM8K test set using different decoding strategies over \texttt{mistral-7b}.
    Nat stands for natural CoT decoding, which is the default decoding strategy in this paper.
    Temp stands for Temperature sampling (\texttt{temperature} = 0.7).
    Beam stands for Beam Search sampling.
  }
\end{table}

\mymodel consistently achieves the best performance across all decoding strategies, delivering an average absolute accuracy improvement of $5.7\%$ compared to Majority Voting. In contrast, CoT-Decoding only achieves only a $1.69\%$ improvement over Majority Voting.
The performance of CoT-Decoding tends to rely more heavily on the format of LLMs' output rather than its correctness.
If the decoding strategy fails to elicit outputs in a specific format (e.g., those containing a CoT process), the effectiveness of CoT-Decoding is compromised. 
In contrast, \mymodel analyzes the correctness of LLMs' outputs and is therefore less affected by changes in sampling strategies compared to CoT-Decoding.




\subsubsection{The effect of hidden states from different layers} 

\normalem{\emph{Hidden States of  middle layer optimize \mymodel’s Performance.}}
Previous research shows that hidden state from the middle to deeper layers contain richer knowledge compared to those in the shallower layers.
\begin{table}[h]
\centering
\begin{tabular}{c|cc|cc}
\toprule
\multirow{2}{*}{\textbf{Layer}} & \multicolumn{2}{c|}{\textbf{GSM8K}} & \multicolumn{2}{c}{\textbf{MMLU-Pro}} \\
                       &\textbf{sum}    & \textbf{max}   & \textbf{sum}   & \textbf{max}            \\ \midrule 
5                      & 51.82          & 36.26          & 41.32          & 38.78          \\
15                     & 51.91          & 37.28          & 41.34          & 39.46 \\
20                     & \textbf{53.14} & \textbf{50.41} & \textbf{41.82} & \textbf{40.36}          \\
25                     & 52.41          & 32.16          & 41.25          & 38.07          \\
32                     & 52.22          & 38.80          & 41.37          & 38.10          \\ \bottomrule
\end{tabular}
\caption{
  \label{tab:result_diff_layers}
    Accuracy of \mymodel on GSM8K test set and MMLU-Pro using  hidden states from different \texttt{mistral-7b} layers.
  }
\end{table}
Table~\ref{tab:result_diff_layers} shows that \mymodel achieves optimal performance when utilizing hidden states from the 15th or 20th layers, which aligns with the conclusions of prior research. 
The selection of hidden states' layer has a significantly greater impact on \mymodel (max) compared to \mymodel (sum). 
For instance, on GSM8K, the standard deviation of accuracy for \mymodel (max) across different layers is $7.1$, 
whereas it is only $0.6$ for \mymodel (sum). 
This is because the knowledge contained in the hidden states directly influences \mymodel (max)'s judgment on the correctness of assertions. 
\mymodel (sum) incorporates the frequency of answers, which mitigates the influence of layer selection.

\subsubsection{The effect of problem difficulty} 
\begin{table}[h]
\centering
\begin{tabular}{c|cc}
\toprule
\textbf{\texttt{max\_op}} & \textbf{Majority Voting} & \textbf{\mymodel}         \\ \midrule
2       & 55.0 & 67.0 (+21.8\%) \\
4       & 40.0 & 47.0 (+17.5\%) \\
8       & 25.0 & 30.0 (+20.0\%) \\
16      & 18.0 & 21.0 (+16.7\%) \\ \bottomrule
\end{tabular}
\caption{
  \label{tab:result_diff_difficulty}
    Accuracy of \mymodel on iGSM with different difficulty over \texttt{mistral-7b}.
    A higher \texttt{max\_op} indicates greater difficulty.
    The values in parentheses indicate the percentage accuracy gain over the baseline.
  }
\end{table}
We investigate the impact of problem complexity using iGSM dataset with varying levels of difficulty.
More detailed information about iGSM dataset is provided in Appendex~\ref{appendix:igsm}.
\normalem{\emph{\mymodel Effectively enhances reasoning accuracy across various levels of problem difficulty.}}
Table~\ref{tab:result_diff_difficulty} demonstrates \mymodel consistently achieves accuracy gains over the baseline across varying levels of problem difficulty, with no significant decline in performance gains as the difficulty increases.
This is attributed to \mymodel's effective utilization of logical rules to harness the internal knowledge of LLMs, suggests that \mymodel exhibits greater robustness in handling complex reasoning problems.

\subsubsection{The effect of numbers of solutions per question} 
\begin{figure}[h]
    \centering
    \includegraphics[width=0.8\linewidth]{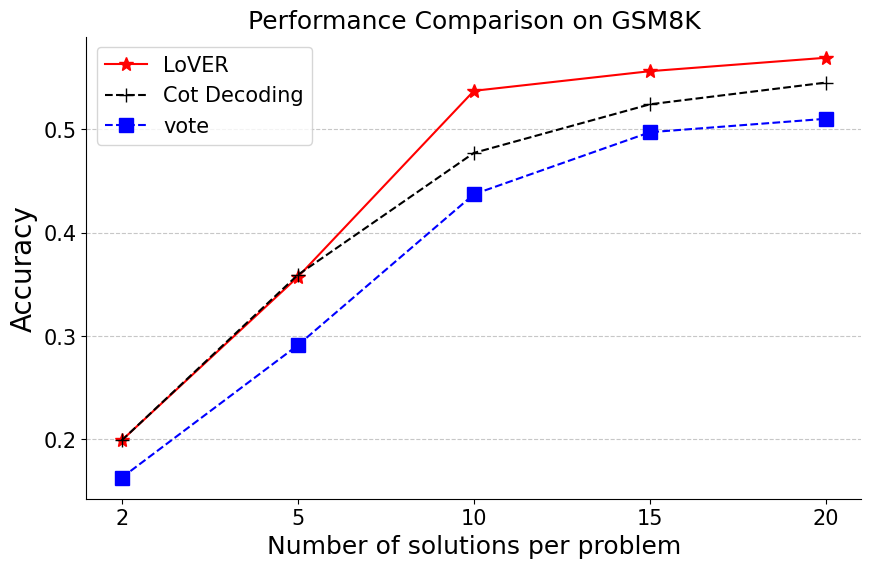}
    \caption{An accuracy comparison of \mymodel and baselines across different numbers of solutions on GSM8K over \texttt{mistral-7b}.}
    \label{fig:result_branch_n}
\end{figure}


\normalem{\emph{\mymodel maintains high reasoning accuracy regardless of $N$.}}
The reasoning accuracy of all methods increases as the number of solutions grows shown in Table \ref{fig:result_branch_n}.
Across different numbers of solutions, \mymodel consistently outperform Majority Voting, achieving an average absolute accuracy gain of $6.4\%$. 
As the number of solutions increases ($>5$), \mymodel surpasses CoT-Decoding, delivering an average absolute accuracy gain of $3.9\%$. 
Compared to CoT-Decoding, \mymodel extracts richer information from hidden states, enabling more effective utilization of the diverse solutions sampled during the decoding process. 


\subsubsection{The effect of model scales} 

\begin{figure}[h]
    \centering
        \includegraphics[width=0.9\linewidth]{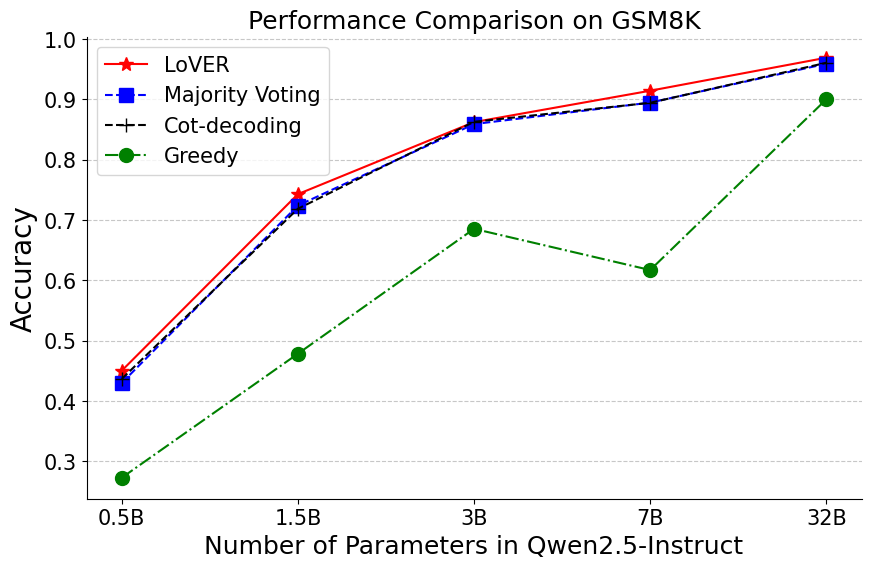}
    \caption{\mymodel reliably improves reasoning performance across model scales (\texttt{Qwen-2.5} family).}
    \label{fig:result_model_scales}
\end{figure}

\normalem{\emph{\mymodel enhances reasoning accuracy across model scales.}
Figure~\ref{fig:result_model_scales} shows that \mymodel enhances reasoning accuracy across different model scales over the \texttt{Qwen-2.5} family.
\mymodel enables a 7B-parameter LLM to achieve reasoning accuracy comparable to that of a 32B-parameter LLM.
\mymodel achieves an average accuracy gains of $1.7\%$ across five models with varying parameter sizes and consistently outperforms the baselines. 
In contrast, CoT-Decoding and Majority Voting exhibit comparable performance.





\subsection{OOD Generalization}

\begin{table}[h]
\centering
\begin{tabular}{l|ccc}
\toprule
\textbf{Datasets}& \textbf{Voting} & \textbf{CoT-D} & \textbf{\mymodel} \\ \midrule
Boolean Exp & 61.6                                & \textbf{72.4}                    & \textbf{\textbf{72.4}}   \\
Web of Lies & 25.6                                & 39.6                             & \textbf{44.8}            \\
Object Cnt  & 46.0                                & 48.0                             & \textbf{48.4}            \\
Navigate    & 49.6                                & 57.2                             & \textbf{57.6}            \\
Arithmetic  & 12.0                                & 11.6                             & \textbf{12.4}            \\
Causal Jud  & 45.9                                & \textbf{60.4}                    & \textbf{60.4}            \\ \bottomrule
\end{tabular}
\caption{
  \label{tab:results_ood}
    Reasoning accuracy of \mymodel on six OOD datasets over \texttt{mistral-7b}. 
    Voting stands for "Majority Voting";
    CoT-D stands for "CoT-Decoding".
    Boolean Exp stands for "Boolean Expression" dataset;
    Object Cnt stands for "Object Counting" dataset;
    Arithmetic stands for "Multi-Step Arithmetic Two" dataset;
    Causal Jud stands for "Causal Judgment" dataset.
  }
\end{table}

\normalem{\emph{The improvements brought by \mymodel  can be transferred to OOD problems.}}
We select six distinct datasets from BIG-Bench Hard as OOD datasets.
As shown in table~\ref{tab:results_ood}, \mymodel achieves higher or equal accuracy than CoT-Decoding across six all datasets, and outperforms Majority Voting on all datasets, with an average accuracy improvement of $2.5\%$. 
This demonstrates the strong OOD generalization capability of \mymodel.
By leveraging the latent knowledge and logical constraints, \mymodel learns patterns for evaluating assertion correctness, independent of the domain-specific context of the assertions.

\section{Related Work}
\label{sec:related}
\paragraph{LLM Reasoning}
Existing research on LLM reasoning can be roughly divided into two categories: extrinsic reasoning and intrinsic reasoning. 
Extrinsic reasoning mainly involves complex prompt engineering~\cite{wei2022chain,yao2024tree}, verifier based on  outcome or processes~\cite{cobbe2021training,lightman2023let}, and customized search algorithms (e.g., A*, MCTS)~\cite{zhuangtoolchain,feng2023alphazero}.
Research in this domain focuses on various reasoning tasks such as mathematical reasoning~\cite{cobbe2021training,DBLP:conf/nips/HendrycksBKABTS21}, logical reasoning~\cite{liulogiqa}, common-sense reasoning~\cite{DBLP:conf/emnlp/Yang0ZBCSM18}, and more~\cite{liu2024investigating,yuan2024analogykb,chen2022kar}. 
Intrinsic reasoning seeks to explore the model’s internal knowledge, primarily through the observation and manipulation of its hidden layers or output probabilities. 
The probe method involves using auxiliary classifiers or probes to analyze and interpret the internal representations learned by the model, offering insights into its understanding and reasoning processes~\cite{belinkov2022probing,alain2018understandingintermediatelayersusing}. 
Unlike supervised probes, CCS~\cite{burns2023discovering} learns a linear classifier to uncover latent knowledge in an unsupervised manner, while CoT-Decoding~\cite{NEURIPS2024_7a8e7fd2} assesses the truth value of candidate solutions based on answer confidence. 
However, the main drawbacks are that CoT-Decoding is essentially an expert-curated heuristic rule, and CCS is limited to Yes-No questions, lacking scalability. 
\mymodel belongs to the second category and is applicable to general reasoning tasks that do not require supervised data. 
It can be seen as an unsupervised probe (verifier) guided by logic rule.

\paragraph{Neural Logical Reasoning}
Neural logic integrates neural networks with logical reasoning to enhance model's interpretability, consistency, and reasoning capabilities. 
One paradigm involves learning logical operators such as AND, OR, and NOT as differentiable neural modules, guided by self-supervised logic regularization \cite{shi2020neural}. 
Prior studies have demonstrated its effectiveness in  proof generation \cite{sun2021probabilistic}, fact checking \cite{chen2022loren}, NLI \cite{li2019logic} and recommender systems \cite{chen2021neural}.
Another standard method is based on the variational EM framework~\cite{ru2021learning,qu2019probabilistic,zhou2020towards}.
\mymodel draws inspiration from both lines of work. 
We represent the output of the verifier as binary latent variables, which are regularized with soft logic.

\section{Conclusion}
\label{sec:conclusion}
We propose \mymodel, an unsupervised verifier regularized by logical rules for enhancing LLMs' reasoning capability.
We design three logical rules to guide \mymodel in effectively leveraging unlabeled data, achieving performance comparable to supervised methods.
\mymodel is compatible with any white-box LLMs and adaptable to diverse reasoning tasks.
Experiments show \mymodel significantly enhances the reasoning accuracy of LLMs while demonstrating strong OOD generalization capabilities. 

\section*{Limitations}
\label{sec:limitations}
\mymodel relies on the hidden states of LLMs, which inherently restricts its applicability to white-box LLMs. 
This dependency prevents \mymodel from being directly utilized in black-box LLMs scenarios.
Although \mymodel does not rely on extracting final answers from responses and can be applied to scenarios where responses lack explicit conclusions, we do not conduct experiments to explore this aspect in this paper.
\section*{Acknowledgments}
The authors wish to thank the reviewers for their helpful
comments and suggestions.
This work was supported by NSFC grant(No.62136002 and 62477014), Ministry of Education Research Joint Fund Project(8091B042239), and Fundamental Research Funds for the Central Universities. 

\bibliography{main}

\clearpage
\newpage
\appendix

\label{sec:appendix}



\section{Logic-based inter-group Consistency}
\label{sec:logic_based}
The basic idea is to transform Eq.~\ref{eq:inter} into a logical expression and then apply product t-norms to relax the logic~\cite{chen2022loren,li2019logic}.
\begin{IEEEeqnarray*}{ccl}
r &= ~  &(\bm{z}_1^+ \wedge \neg \bm{z}_2^+ \wedge \dots \wedge \neg \bm{z}_M^+) \\
& \vee & (\neg \bm{z}_1^+ \wedge \bm{z}_2^+ \wedge \dots \wedge \neg \bm{z}_M^+) \\
& \vdots & \\
& \vee &(\neg \bm{z}_1^+  \dots \wedge \bm{z}_M^+), \\
\mathcal{L}_\mathrm{inter}^\mathrm{logic} &=&  \mathtt{t}\text{-}\mathtt{norms}(r).
\end{IEEEeqnarray*}

The accuracy comparison of \mymodel using using different kinds of $\mathcal{L}_\mathrm{inter}$ are provided in Table~\ref{tab:res_logic}.
\mymodel using default $\mathcal{L}_\mathrm{inter}$ is better.

\begin{table}[h]
\centering
\begin{tabular}{l|ll}
\hline
        & \multicolumn{2}{l}{\textbf{GSM8K}} \\
        & \textbf{sum}     & \textbf{max}    \\ \hline
default $\mathcal{L}_\mathrm{inter}$ & \textbf{53.14}            & \textbf{50.41}           \\
logic-based $\mathcal{L}_\mathrm{inter}$   & 52.99            & 46.39           \\ \hline
\end{tabular}
\caption{
\label{tab:res_logic}
Reasoning accuracy of \mymodel on GSM8K test set over \texttt{mistral-7b} using different kinds of $\mathcal{L}_\mathrm{inter}$. 
}
\end{table}

\section{Datasets}
\label{sec:datasets}
\subsection{HotpotQA}

HotpotQA is a Wikipedia-based question-answering dataset. In the full wiki setting, HotpotQA consists of 90,447 training samples, 7,405 validation samples, and 7,405 test samples. 
A key feature of HotpotQA is that these questions require finding and reasoning over multiple supporting documents to answer.

Since the test set does not provide standard answers, we use the validation set as the test set.
To make the extraction of final answers from response easier and more accurate, we only selected questions with "yes" or "no" as answers for our experiments. 
This resulted in 5,481 training samples and 458 test samples. 
Table~\ref{tab:data_hotpot_example} provides examples of questions with "yes" and "no" answer.
We used the entire test set and a randomly selected subset of 3,000 training samples for the experiments.

In our experiments, we only utilized the questions and did not incorporate the related supporting texts.

\begin{table}[h]
\centering
\begin{tabular}{p{0.96\linewidth}}
\toprule
\textbf{HotpotQA}
\\
\midrule
\textbf{Question}: Were Scott Derrickson and Ed Wood of the same nationality?
\\
\textbf{Answer}: yes
\\
\midrule
\textbf{Question}: Were Pavel Urysohn and Leonid Levin known for the same type of work?
\\
\textbf{Answer}: no
\\
\bottomrule
\end{tabular}
\caption{
\label{tab:data_hotpot_example}
Examples of HotpotQA dataset with "yes" and "no" answer.
}
\end{table}

\subsection{MMLU pro}

MMLU-Pro is an advanced benchmark for assessing language models on more extensive and challenging tasks.
It consists of over 12,000 questions, each furnished with ten possible answers, spanning 14 distinct domains. 
These domains include Biology, Business, Chemistry, Computer Science, Economics, Engineering, Health, History, Law, Math, Philosophy, Physics, Psychology, and Others.

MMLU-Pro provides over 12,000 test samples and 70 validation samples, with no training set included. 
Therefore, we randomly selected one-third (4 out of 14) of the domains to serve as the test set, while the remaining domains were used for training and validation. 
With the random seed set to 0, we obtained data from the following four domains as the test set: Computer Science, Physics, History, and Biology.
The statistics of test set is shown in Tabel~\ref{tab:data_mmlu_test}.

\begin{table}[h]
\centering
\begin{tabular}{lc}
\toprule
\textbf{Datasets}         & \textbf{\#Data} \\ \hline
Computer Science & 410    \\
Physics          & 1299   \\
History          & 381    \\
Biology          & 717    \\ \hline
All              & 2807   \\ \bottomrule
\end{tabular}
\caption{
\label{tab:data_mmlu_test}
The statistics of the four test datasets from MMLU-Pro.
}
\end{table}

\subsection{BIG-Bench Hard}
\label{appendix:bbh}

We randomly selected six datasets from BIG-Bench Hard as out-of-distribution (OOD) datasets, with the specific dataset names and corresponding statistics summarized in Table~\ref{tab:data_bigbench}.

\begin{table}[h]
\centering
\begin{tabular}{lc}
\toprule
\textbf{Datasets}                 & \textbf{\#Data} \\ \hline
Boolean Expression        & 250    \\
Web of Lies               & 250    \\
Object Counting           & 250    \\
Navigate                  & 250    \\
Multi-Step Arithmetic Two & 250    \\
Causal Judgment           & 187    \\ \hline
All                       & 1437   \\ \bottomrule
\end{tabular}
\caption{
\label{tab:data_bigbench}
The statistics of the six OOD datasets from BIG-Bench Hard.
}
\end{table}

\subsection{iGSM}
\label{appendix:igsm}

We construct a more challenging mathematics dataset called iGSM.
Following iGSM\cite{YXLA2024-gsm1}, we control problems difficulty by setting different numbers of operations (\textbf{max\_op}) in the solutions. 
We utilized the iGSM synthetic data generator\footnote{https://github.com/facebookresearch/iGSM} to construct the dataset by setting \verb|random_seed| to 42, \verb|max_edge| to 12, \verb|perm_level| to 5, \verb|detail_level| to 0, and \verb|max_op| to 2, 4, 8, and 16, respectively. The detailed data statistics are presented in Table ~\ref{tab:data_igsm_test}.

\begin{table}[h]
\centering
\begin{tabular}{lcc}
\toprule
\multirow{2}{*}{\textbf{max\_op}} & \multicolumn{2}{c}{\textbf{\#Data}} \\
                         & \textbf{test}       & \textbf{training}      \\ \midrule
2                        & 100        & -             \\
4                        & 100        & -             \\
8                        & 100        & 1500          \\
16                       & 100        & -             \\ \hline
All                      & 400        & 1500          \\ \bottomrule
\end{tabular}
\caption{
\label{tab:data_igsm_test}
The statistics of the iGSM test set in our experiments.
}
\end{table}

Table~\ref{tab:data_igsm_example} provides two examples of iGSM where \verb|max_op| is 2 and 16 respectively.

\begin{table}[h]
\begin{tabular}{p{\linewidth}}
\toprule
\textbf{iGSM: \texttt{max\_op=2}}
\\
\midrule

\textbf{Question}: The number of each Goldfish's
Proximal Phalanx equals 8. The number of
each Swordfish's Metacarpal I equals the
sum of each Goldfish's Bone, each Goldfish's
Proximal Phalanx and each Eel's Bone. The
number of each Mahi Mahi's Radial Carpal
equals 3. The number of each Goldfish's
Metacarpal IV equals 12. How many Radial
Carpal does Mahi Mahi have?
\\
\textbf{Solution}: Define Mahi Mahi's Radial Carpal
as n; so n = 3.
\\
\textbf{Answer}: 3
\\
\bottomrule
\end{tabular}
\caption{
\label{tab:data_igsm_example}
Examples of iGSM dataset with different \texttt{max\_op}.
}
\end{table}

\begin{table}[h]
\begin{tabular}{p{\linewidth}}
\toprule
\textbf{iGSM: \texttt{max\_op=16}}
\\
\midrule

\textbf{Question}: The number of each Spinal Cord's
Transitional Epithelial Cells equals the
difference of each Cerebellum's Hepatocytes and
each Albatross's Spinal Cord. The number of each Cerebellum's
Hepatocytes equals each Parrot's Organs. The number of
each Albatross's Spinal Cord equals each Cerebellum's
Hepatocytes. The number of each Spinal Cord's Hepatocytes
equals the sum of each Cerebellum's Hepatocytes, each
Albatross's Spinal Cord and each Parrot's Cerebellum.
The number of each Parrot's Cerebellum equals 6. How
many Cells does Albatross have?
\\

\textbf{Solution}: Define Parrot's Cerebellum as c;
so c = 6. Define Parrot's Organs as i; so i = c = 6.
Define Cerebellum's Hepatocytes as X; so X = i = 6.
Define Albatross's Spinal Cord as u; so u = X = 6.
Define Spinal Cord's Hepatocytes as S;
b = X + u = 6 + 6 = 12; so S = b + c = 12 + 6 = 18.
Define Spinal Cord's Transitional Epithelial Cells as F;\\
so F = X - u = 6 - 6 = 0. Define Spinal Cord's Cells as W;
so W = S + F = 18 + 0 = 18. Define Albatross's Cells as o;
so o = u * W = 6 * 18 = 16., 
\\
\textbf{Answer}: 16
\\
\bottomrule
\end{tabular}
\caption{
\label{tab:data_igsm_example}
Examples of iGSM dataset with different \texttt{max\_op}.
}
\end{table}

\subsection{GSM8K}
GSM8K consists of 8,792 high quality grade school math problems, with 7,473 in the training set and 1,319 in the test set.
GSM8K is designed to evaluate the mathematical reasoning capabilities of models.
In our experiments, we select 7,000 samples randomly from the training set for training, 473 samples from the training set as the validation set, and the entire test set for testing.

\newpage
\section{Pseudocode (PyTorch-like)}
\label{sec:pseudo}

\begin{lstlisting}[language=Python,frame=shadowbox]
"""Step 1: Sample n responses from LLM"""
def sample_responses(llm, prompt, n):
    responses = []
    
    for _ in range(n):
        response = model.sample(prompt)
        responses.append(response)
    return responses

"""Step 2: Create contrastive assertions"""
def create_assertions(question, responses):
    pos_assertions = []
    neg_assertions = []
    
    for response in responses:
        qa_pair = f"Q:{question} A:{response}"
        
        pos_assertion= qa_pair+"this is a true answer."
        neg_assertion= qa_pair+"this is a false answer."
        
        pos_assertions.append(positive_assertion)
        neg_assertions.append(negative_assertion)
    
    return pos_assertions, neg_assertions

"""Step 3: Get hidden states of assertions"""
def get_hidden_states(llm, assertions, layer_idx):
    representations = []
    
    for assertion in assertions:
        outputs = llm(assertion, output_hidden_states)
        hidden_state = outputs.hidden_states[layer_idx]
        hidden_state = hidden_state[:, -1, :]
        representations.append(hidden_state)
    
    return representations

"""Step 4: Compute scores using verifier"""
def compute_scores(verifier, pos_reps, neg_reps):
    pos_scores = verifier(pos_reps)
    neg_scores = verifier(neg_reps)
    
    final_scores = 0.5 * (pos_scores+(1-neg_scores))
    
    return final_scores

"""Step 5a: Select final answer by Max strategy"""
def select_by_max(responses, scores):
    max_idx = argmax(scores)
    best_response = responses[max_idx]
    
    return extract_final_answer(best_response)

"""Step 5b: Select final answer by Sum strategy"""
def select_by_sum(responses, scores)
    answer_scores = {}
    
    for response, score in zip(responses, scores):
        final_answer = extract_final_answer(response)
        answer_scores[final_answer] += score
    
    best_answer = find_key_by_max_value(answer_scores)
    return best_answer
\end{lstlisting}

\newpage
\begin{lstlisting}[language=Python,frame=shadowbox]
if __name__ == "__main__":
    # Step 1: Generate responses
    prompt = construct_prompt(q)
    responses = sample_responses(llm, prompt, n)

    # Step 2: Create contrastive assertions
    pos_asserts, neg_asserts=create_assertions(q, responses)
    
    # Step 3: Get hidden representations
    layer = 20
    pos_reps = get_hidden_states(llm, pos_asserts, layer)
    neg_reps = get_hidden_states(llm, neg_asserts, layer)
    
    # Step 4: Compute verification scores
    scores = compute_scores(verifier, pos_reps, neg_reps)
    
    # Step 5: Select final answer based on strategy
    final_answer_max=select_by_max(responses, scores)
    final_answer_sum=select_by_sum(responses, scores)


\end{lstlisting}

\end{document}